\documentclass[11pt]{article}

\usepackage[preprint]{acl}

\usepackage{times}
\usepackage{latexsym}

\usepackage[T1]{fontenc}
\usepackage[utf8]{inputenc}

\usepackage{microtype}

\usepackage{inconsolata}

\usepackage{graphicx}

\makeatletter
\let\cas@beginabstract\abstract      % \abstract is the *begin* macro
\let\cas@endabstract  \endabstract   % \endabstract is the *end*  macro
\makeatother
\usepackage{arabtex}
\usepackage{utf8}
\setcode{utf8}

\makeatletter
\let\abstract   \cas@beginabstract   % restore begin
\let\endabstract\cas@endabstract     % restore end
\makeatother
\title{AraModernBERT: Transtokenized Initialization and Long-Context Encoder Modeling for Arabic}

\author{
  Omar Elshehy\textsuperscript{1,7} \quad
  Omer Nacar\textsuperscript{2,7} \quad
  Abdelbasset Djamai\textsuperscript{3,7} \quad
  Muhammed Ragab\textsuperscript{4,7} \\
  \textbf{Khloud Al Jallad}\textsuperscript{5,7} \quad
  \textbf{Mona Abdelazim}\textsuperscript{6,7} \\
  \\
  \textsuperscript{1}Universität des Saarlandes,
  \textsuperscript{2}Tuwaiq Academy,
  \textsuperscript{3}Datategy, \\
  \textsuperscript{4}Leibniz-Institute for Educational Media | Georg-Eckert-Institute, \\
  \textsuperscript{5}Arab International University,
  \textsuperscript{6}Ain Shams University,
  \textsuperscript{7}NAMAA Community \\
  \texttt{o.najar@tuwaiq.edu.sa}
}

\begin{document}
\maketitle

\begin{abstract}
Encoder-only transformer models remain widely used for discriminative NLP tasks, yet recent architectural advances have largely focused on English. In this work, we present \textbf{AraModernBERT}\footnote{\url{https://huggingface.co/NAMAA-Space/AraModernBert-Base-V1.0}}, an adaptation of the ModernBERT encoder architecture to Arabic, and study the impact of transtokenized embedding initialization and native long-context modeling up to 8,192 tokens. We show that transtokenization is essential for Arabic language modeling, yielding dramatic improvements in masked language modeling performance compared to non-transtokenized initialization. We further demonstrate that AraModernBERT supports stable and effective long-context modeling, achieving improved intrinsic language modeling performance at extended sequence lengths. Downstream evaluations on Arabic natural language understanding tasks, including inference, offensive language detection, question-question similarity, and named entity recognition, confirm strong transfer to discriminative and sequence labeling settings. Our results highlight practical considerations for adapting modern encoder architectures to Arabic and other languages written in Arabic-derived scripts.
\end{abstract}

% Introduction
\section{Introduction}

Transformer-based encoder-only language models such as BERT have become essential components of modern natural language processing (NLP) pipelines, especially for retrieval, classification, and representation learning tasks \citep{devries2021asgoodasnew,karpukhin2020dpr,khattab2020colbert}. Despite the recent dominance of large autoregressive language models, encoder-based architectures remain widely deployed due to their favorable trade-offs in efficiency, latency, and scalability. Recent work has significantly modernized encoder architectures through improved attention mechanisms, positional encodings, and hardware-aware design, leading to substantial gains in performance and efficiency \citep{warner2025smarter}. However, these advances have been developed and evaluated primarily for English, and their transfer to Arabic and other languages using the Arabic script remains comparatively underexplored.

Arabic presents distinct challenges for encoder-based modeling. Its rich and templatic morphology, high lexical sparsity, and orthographic variation amplify the importance of tokenizer design and embedding initialization strategies \citep{rust2021tokenizer,petrov2023unfair}. Multilingual and English-centric tokenizers often fragment Arabic words excessively, resulting in longer effective sequence lengths and poorly trained subword embeddings. These issues are further compounded in Arabic-language domains such as news, legal texts, religious writings, and encyclopedic content, where documents frequently exceed the 512-token context limit of classical BERT-style models \citep{antoun2020arabert,abdulmageed2021arbert,inoue2021camelbert}. As a result, both tokenization quality and long-context modeling are particularly important for Arabic, yet their interaction with modern encoder architectures has not been systematically studied.

In this paper, we introduce \textbf{AraModernBERT}, an Arabic adaptation of the ModernBERT encoder architecture \citep{warner2025smarter}. Rather than proposing a new model family, we focus on carefully transferring a modernized encoder design to Arabic and empirically analyzing two key factors: \emph{transtokenized embedding initialization} and \emph{native long-context modeling up to 8{,}192 tokens}. Transtokenization aligns a newly trained tokenizer with pretrained representations by initializing target-language embeddings from semantically aligned source-language embeddings, thereby mitigating the mismatch between tokenizer vocabularies and embedding spaces \citep{remy2024trans}. Long-context modeling, enabled by architectural design choices such as alternating local and global attention and rotary positional embeddings \citep{su2021roformer}, allows the encoder to process substantially longer sequences than traditional Arabic BERT variants.

We conduct a comprehensive evaluation spanning intrinsic language modeling, downstream Arabic natural language understanding (NLU) tasks, and retrieval. Our experiments show that transtokenization is essential for stable and effective Arabic encoder training, achieving considerable improvements in masked language modeling performance compared to non-transtokenized initialization. We further demonstrate that AraModernBERT supports stable long-context modeling, achieving improved masked language modeling performance at extended sequence lengths without numerical instability or excessive memory usage. Downstream evaluations on Arabic natural language understanding tasks, including natural language inference (NLI), offensive language detection, and question--question similarity, confirm strong transfer to discriminative settings \citep{antoun2020arabert,abdulmageed2021arbert}.

This work provides practical insights into adapting modern encoder architectures to Arabic. By focusing on tokenizer initialization and long-context modeling, we highlight design considerations that are broadly applicable to Arabic and other Arabic-script languages. We release AraModernBERT and our evaluation code to support further research in this space.

\section{Related Work}

Encoder-only transformer models have been widely adopted for Arabic NLP, with AraBERT and its variants establishing strong baselines for Modern Standard Arabic and selected dialects \citep{antoun2020arabert}. Subsequent work, including CAMeLBERT and MARBERT, demonstrated the importance of domain selection and dialectal coverage for Arabic pretraining \citep{inoue2021camelbert,abdulmageed2021arbert}. Despite their effectiveness, these models largely inherit the original BERT design, including a fixed 512-token context limit and absolute positional embeddings, which restrict their applicability to long Arabic documents commonly found in news, legal, and religious domains.

While some recent Arabic encoder efforts focus on efficiency or specialization, architectural modernization has largely lagged behind advances developed for English-language encoders. In contrast, a growing body of work revisits encoder design more broadly. Models such as MosaicBERT, AcademicBERT, and CrammingBERT explore training efficiency and resource-constrained settings, but do not substantially alter core architectural assumptions such as context length or attention structure. More recent long-context encoders, including NomicBERT and GTE-en-MLM, extend sequence length primarily for retrieval-oriented applications, but are trained and evaluated almost exclusively on English, limiting their relevance to morphologically rich and under-resourced languages.

ModernBERT represents a significant step forward in encoder architecture by incorporating alternating local and global attention, rotary positional embeddings, and hardware-aware design, enabling native processing of sequences up to 8{,}192 tokens while maintaining high efficiency \citep{warner2025smarter}. Our work builds directly on this architecture and investigates its transfer to Arabic, a setting not explored in the original ModernBERT study.

Tokenization has been shown to play a central role in multilingual and low-resource language modeling. Prior work demonstrates that multilingual subword tokenizers disproportionately benefit high-resource languages with shared alphabets, often leading to excessive fragmentation and poorly trained embeddings for languages such as Arabic \citep{rust2021tokenizer,petrov2023unfair}. Vocabulary transfer has therefore emerged as a promising strategy for language adaptation, with early approaches relying on embedding alignment or token reuse based on orthographic similarity \citep{artetxe2020crosslingual,devries2021asgoodasnew}. However, these methods are limited by tokenizer overlap and language proximity.

\begin{figure*}[t]
\centering
\includegraphics[width=\linewidth]{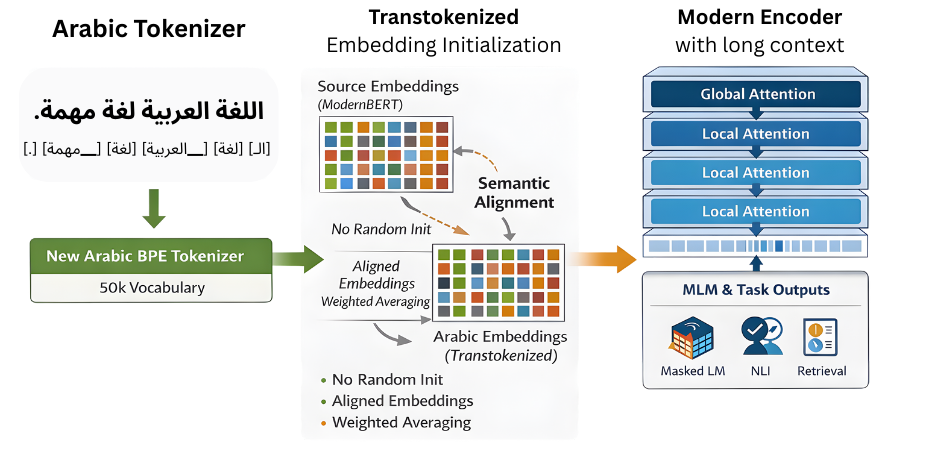}
\caption{AraModernBERT integrates an Arabic BPE tokenizer with transtokenized embedding initialization and a ModernBERT encoder supporting native long-context modeling up to 8{,}192 tokens.}
\label{fig:aramodernbert}
\end{figure*}

Trans-tokenization addresses these limitations by explicitly aligning token vocabularies using parallel corpora and statistical alignment, initializing target-language embeddings as weighted combinations of semantically aligned source embeddings \citep{remy2024trans}. This approach has been shown to enable stable adaptation of large language models to low-resource languages without catastrophic degradation. In contrast to prior work focusing on cross-lingual transfer for generative models, we adopt transtokenization in a monolingual Arabic setting and demonstrate its critical role in training modern Arabic encoder models.

Finally, long-context modeling and retrieval have received increasing attention as NLP applications move toward document-level understanding. While extended context improves language modeling capacity, prior work shows that naïvely encoding long documents into a single vector often degrades retrieval performance due to representation dilution, motivating multi-vector and late-interaction approaches such as ColBERT \citep{khattab2020colbert,karpukhin2020dpr}. In Arabic NLP, long-context retrieval remains underexplored, and most systems rely on chunking long documents to fit short-context encoders. Our work contributes empirical evidence to this discussion by analyzing long-context retrieval with a modern Arabic encoder and clarifying when architectural changes beyond context length are required.

\section{Methodology}

This section describes the design and training of \textbf{AraModernBERT}, an Arabic encoder model adapted from the ModernBERT architecture. Our methodology focuses on two central aspects: (i) the transfer of a modernized encoder architecture to Arabic and (ii) the use of transtokenized embedding initialization to enable stable and effective Arabic language modeling. Figure~\ref{fig:aramodernbert} provides an overview of the full pipeline, illustrating how a new Arabic tokenizer is introduced, how its embeddings are initialized via transtokenization, and how the resulting representations are processed by a long-context encoder.

Concretely, given a new Arabic tokenizer and a pretrained source embedding space, transtokenization proceeds by aligning target-language tokens to semantically related source-language tokens using a parallel corpus and statistical alignment. For each Arabic token $t$, we obtain a set of aligned source tokens $\{s_i\}$ with associated alignment counts $c_{t \rightarrow s_i}$. The embedding of $t$ is then initialized as a weighted average of the aligned source embeddings:

\begin{equation}
\mathbf{e}(t) \;=\; \sum_{i} \frac{c_{t \rightarrow s_i}}{\sum_{j} c_{t \rightarrow s_j}} \, \mathbf{e}(s_i),
\label{eq:transtokenization}
\end{equation}

where $\mathbf{e}(s_i)$ denotes the pretrained embedding of source token $s_i$, and $c_{t \rightarrow s_i}$ is the alignment count between target token $t$ and source token $s_i$. The normalization ensures that the weights form a probability distribution over aligned source tokens.

For example, the Arabic token \RL{اللغة} may align to English tokens such as \textit{language} and \textit{linguistic}, and its embedding is initialized as the normalized weighted combination of the corresponding source embeddings. Tokens without reliable alignments are initialized using predefined fallback mappings (e.g., digits, punctuation, or special symbols). This procedure avoids random initialization while preserving semantic structure in the embedding space.

As shown in Figure~\ref{fig:aramodernbert}, transtokenization injects semantically aligned pretrained embeddings into the newly introduced Arabic tokenizer, avoiding the performance degradation typically caused by random embedding initialization. This step is critical for stable masked language model training in Arabic and allows the encoder to fully benefit from the modern architectural features of ModernBERT, including long-context processing.

AraModernBERT is an encoder-only transformer model built on top of the ModernBERT architecture. We retain all core architectural design choices of ModernBERT, which were originally proposed to address efficiency and scalability limitations of classical BERT-style encoders. In particular, AraModernBERT employs a stack of 22 transformer layers with a hidden dimension of 768 and 12 attention heads, resulting in approximately 149 million parameters.

A key feature of the architecture is its \emph{alternating attention mechanism}. Every third layer applies global self-attention, allowing tokens to attend to the entire sequence, while the remaining layers use local self-attention with a sliding window of 128 tokens. This design balances long-range dependency modeling with computational efficiency and enables native processing of long documents.

\paragraph{Context Modeling.} AraModernBERT natively supports a maximum sequence length of 8{,}192 tokens. Long-context capability is enabled through the use of Rotary Positional Embeddings (RoPE), with distinct configuration parameters for global and local attention layers. Specifically, global attention layers use a RoPE theta value of 160{,}000, while local attention layers use a theta of 10{,}000. This separation allows the model to maintain positional sensitivity across both short- and long-range interactions.

Importantly, long-context modeling in AraModernBERT is \emph{native} rather than windowed: the full sequence is processed in a single forward pass without truncation or recurrence. This design is particularly well-suited to Arabic-language domains where documents frequently exceed the 512-token limit of traditional encoders.

\paragraph{Arabic Tokenization.} Given the morphological richness and orthographic characteristics of Arabic, we train a dedicated Arabic tokenizer rather than reusing multilingual or English-centric tokenizers. The tokenizer is based on byte-pair encoding (BPE) and has a vocabulary size of 50{,}280 tokens, optimized to capture common Arabic morphemes and word forms while reducing excessive subword fragmentation.

Special tokens follow standard encoder conventions, including dedicated tokens for classification, masking, padding, and separation. This tokenizer serves as the foundation for all pretraining and downstream evaluation.

\paragraph{Transtokenized Embedding Initialization.} Replacing a tokenizer in a pretrained model typically requires reinitializing the embedding table, which can lead to severe degradation in performance. To address this issue, AraModernBERT adopts the \emph{transtokenization} strategy for embedding initialization. Transtokenization initializes the embedding vectors of the new Arabic tokenizer using a weighted combination of semantically aligned embeddings from a source model, rather than random initialization. This alignment is derived from cross-lingual token mappings based on translation resources and statistical alignment techniques. By preserving semantic structure in the embedding space, transtokenization enables stable training and effective transfer even when introducing a new tokenizer.

In AraModernBERT, transtokenization is applied to the input embedding layer prior to masked language model training. Our ablation experiments demonstrate that this step is essential for successful Arabic encoder training.

\paragraph{Training Objective and Data.} AraModernBERT is trained using the masked language modeling (MLM) objective. During training, 30\% of input tokens are masked following standard MLM procedures. Pretraining is conducted on approximately 100 gigabytes of Arabic text drawn from diverse sources, covering a range of domains and writing styles.

Training proceeds in two stages. The model is first trained at shorter sequence lengths to establish stable representations, and subsequently trained with extended sequences up to 8{,}192 tokens to enable long-context modeling. No task-specific supervision is used during pretraining. Table~\ref{tab:config} summarizes the key architectural and training parameters of AraModernBERT.

\begin{table}[h]
% \centering
\resizebox{\columnwidth}{!}{%
\begin{tabular}{ll}
\hline
\textbf{Parameter} & \textbf{Value} \\
\hline
Architecture & ModernBERT encoder \\
Hidden size & 768 \\
Transformer layers & 22 \\
Attention heads & 12 \\
Intermediate size & 1{,}152 \\
Vocabulary size & 50{,}280 \\
Maximum context length & 8{,}192 \\
Global attention frequency & Every 3 layers \\
Local attention window & 128 tokens \\
RoPE theta (global) & 160{,}000 \\
RoPE theta (local) & 10{,}000 \\
Training objective & MLM \\
\hline
\end{tabular}
}
\caption{AraModernBERT configuration and architectural parameters.}
\label{tab:config}
\end{table}

\section{Experiments and Results}

\begin{table*}[t]
\centering
\begin{tabular}{lcc}
\hline
\textbf{Model Variant} & \textbf{MLM Loss} $\downarrow$ & \textbf{Perplexity} $\downarrow$ \\
\hline
AraModernBERT (Transtokenized) & \textbf{3.24} & \textbf{25.54} \\
Embedding Re-initialized & 11.46 & 94{,}372 \\
Fully Random Initialization & 10.98 & 58{,}962 \\
\hline
\end{tabular}
\caption{Transtokenization ablation results on Arabic MLM.}

\label{tab:transtokenization}
\end{table*}

This section presents an empirical evaluation of AraModernBERT across intrinsic language modeling, downstream Arabic natural language understanding, and retrieval. Our experiments are designed to assess three core aspects: (i) the impact of transtokenized embedding initialization, (ii) the effectiveness of native long-context modeling, and (iii) the extent to which the learned representations transfer to downstream Arabic tasks.

\subsection{Experimental Setup}

We conduct intrinsic evaluations using masked language modeling (MLM) on Arabic Wikipedia. Downstream tasks are evaluated by fine-tuning AraModernBERT with task-specific classification heads on top of the encoder, following standard training protocols. For retrieval, we adopt a dense bi-encoder setup with cosine similarity and in-batch negatives where applicable. All experiments are performed with fixed random seeds and consistent hyperparameter settings to ensure reproducibility.

\subsection{Evaluation Metrics}

We adopt standard evaluation metrics appropriate for each task. For intrinsic language modeling, we report MLM loss and perplexity, where lower values indicate better modeling performance. For downstream Arabic natural language understanding tasks, we use accuracy for natural language inference and macro-averaged F1 score for classification tasks with class imbalance, including offensive language detection and question-question similarity. For retrieval experiments, we report Recall@k (with $k \in \{1,5,10\}$) and Mean Reciprocal Rank (MRR), which measure the ability of the model to rank relevant documents highly. These metrics are widely used in prior work and provide complementary perspectives on model effectiveness across tasks.

\subsection{Intrinsic Evaluation: Transtokenization Ablation}

To isolate the effect of transtokenized embedding initialization, we compare AraModernBERT against two ablated variants: (i) an embedding re-initialized model, where the tokenizer is kept fixed but the embedding table is randomly reinitialized, and (ii) a fully randomly initialized model with the same architecture.

The results as shown in Table~\ref{tab:transtokenization}, show that transtokenization is critical for Arabic encoder training. Reinitializing the embedding table leads to catastrophic degradation, increasing perplexity by several orders of magnitude. This confirms that embedding initialization plays a central role in stabilizing Arabic language modeling when introducing a new tokenizer.

\subsection{Long-Context Language Modeling}

We evaluate AraModernBERT under its native 8{,}192-token context by concatenating Arabic Wikipedia articles into long sequences and computing MLM loss. For comparison, we also report performance at the standard 512-token context length.

\begin{table}[h]
% \centering
\resizebox{\columnwidth}{!}{%
\begin{tabular}{lcc}
\hline
\textbf{Context Length} & \textbf{MLM Loss} $\downarrow$ & \textbf{Perplexity} $\downarrow$ \\
\hline
512 tokens & 3.24 & 25.54 \\
8{,}192 tokens & \textbf{3.05} & \textbf{21.05} \\
\hline
\end{tabular}
}
\caption{Masked language modeling performance at different context lengths.}
\label{tab:longcontext}
\end{table}

Interestingly, as shown in Table \ref{tab:longcontext}, MLM loss and perplexity improve at extended context lengths. This indicates that AraModernBERT effectively exploits long-range contextual information rather than suffering from instability or degradation. The model remains memory-efficient, requiring approximately 6.8 GB of GPU memory for 8k-token inference.

\subsection{Arabic Natural Language Understanding}

We evaluate AraModernBERT on three representative Arabic natural language understanding (NLU) tasks: natural language inference, toxicity detection, and semantic similarity. We use the Arabic subset of XNLI \citep{conneau2018xnli}, the OSACT4 Offensive Language Detection (OOLD) dataset \citep{mubarak-etal-2020-overview}, and the Mawdoo3 Question Semantic Similarity (MQ2Q) dataset \citep{seelawi-etal-2019-nsurl}. For computational consistency, all reported results are obtained on fixed test subsets of 2{,}000 instances per task. Each task is fine-tuned using a standard classification head on top of the encoder.

\begin{table}[h]
\centering
\resizebox{\columnwidth}{!}{%
\begin{tabular}{lcc}
\hline
\textbf{Task} & \textbf{Metric} & \textbf{AraModernBERT} \\
\hline
XNLI (Arabic) & Accuracy & 0.47 \\
OOLD & F1-macro & 0.87 \\
MQ2Q  & F1-macro & 0.96 \\
\hline
\end{tabular}
}
\caption{Arabic natural language understanding results.}
\label{tab:nlu}
\end{table}

As shown in Table~\ref{tab:nlu}, AraModernBERT demonstrates strong transfer to downstream Arabic NLU tasks, particularly for semantic similarity and offensive language detection. Performance on Arabic XNLI is consistent with prior encoder-based models and reflects the limited size and label noise of available Arabic NLI resources.

\subsection{Arabic Retrieval}

\paragraph{Short-Text Retrieval.}
We evaluate short-text semantic retrieval using MQ2Q in a dense bi-encoder setting. Questions are treated as queries and their paired equivalents as relevant documents. AraModernBERT is compared against a representative Arabic encoder baseline, AraBERT-base, under identical training and evaluation conditions.

\begin{table*}[h]
\centering
\begin{tabular}{lcccc}
\hline
\textbf{Model} & R@1 & R@5 & R@10 & MRR \\
\hline
AraBERT-base & 0.54 & 0.97 & 0.99 & 0.73 \\
AraModernBERT & 0.52 & 0.97 & 0.99 & 0.72 \\
\hline
\end{tabular}
\caption{Short-text retrieval results on MQ2Q.}
\label{tab:retrieval-short}
\end{table*}

As shown in Table~\ref{tab:retrieval-short}, both models achieve strong retrieval performance. AraBERT slightly outperforms AraModernBERT in this setting, which favors short, lexically similar queries. This result indicates that AraModernBERT remains competitive for short-text semantic retrieval, while its primary advantages lie in representation learning and long-context modeling rather than lexical matching.

\subsection{Arabic Named Entity Recognition}

Named Entity Recognition (NER) has long been a core task in Arabic NLP, with early systems relying on statistical and rule-based methods tailored to Arabic morphology and orthography \citep{benajibaANERsysArabicNamed2007}. Subsequent work explored the use of cross-lingual resources and multilingual transfer to mitigate data sparsity in Arabic NER \citep{darwishNamedEntityRecognition2013,rahimiMassivelyMultilingualTransfer2019}. More recent neural approaches have demonstrated strong performance on Arabic NER when sufficient annotated data and appropriate pretraining are available, though performance remains sensitive to domain, noise, and sentence structure \citep{schneiderCoarseLexicalSemantic2012}.

We further evaluate AraModernBERT on Arabic named entity recognition to assess its effectiveness on sequence labeling tasks. Experiments are conducted on multiple Arabic NER benchmarks, including WikiAnn (Arabic) \citep{rahimiMassivelyMultilingualTransfer2019}, ANERCorp \cite{benajibaANERsysArabicNamed2007}, and AQMAR~\cite{mohit2012recall}. All models use a standard token-level classification head and are evaluated using entity-level F1 score, with results averaged over three random seeds.

\begin{table}[t]
\centering
\begin{tabular}{lcc}
\hline
\textbf{Dataset} & \textbf{Validation F1} & \textbf{Test F1} \\
\hline
WikiAnn (ar) & 0.8571 & 0.8576 \\
ANERCorp & 0.8065 & 0.6827 \\
AQMAR & 0.5541 & 0.5929 \\
Twitter NER & 0.5529 & 0.4919 \\
\hline
\end{tabular}
\caption{Arabic NER results for AraModernBERT. Scores are entity-level F1 averaged over three seeds.}
\label{tab:ner}
\end{table}

AraModernBERT achieves its strongest performance on WikiAnn as shown in Table~\ref{tab:ner}, a large-scale and relatively clean NER benchmark with longer average sentence lengths and substantial training data. Performance is more moderate on smaller or noisier datasets such as ANERCorp, AQMAR, and Twitter NER, which include shorter sentences, higher lexical variability, and domain-specific noise. This pattern suggests that AraModernBERT benefits most from settings where richer sentence-level context and larger annotated corpora align with its pretraining regime on long-form, well-structured Arabic text. Similar trends have been observed in prior Arabic NER studies, where encoder-based models trained on clean data exhibit reduced robustness on noisy or informal text \citep{darwishNamedEntityRecognition2013,rahimiMassivelyMultilingualTransfer2019}.

Across experiments, we find that transtokenization is essential for stable Arabic encoder training and that native long-context modeling improves intrinsic language modeling performance. AraModernBERT transfers effectively to downstream Arabic tasks, including natural language understanding, short-text retrieval, and named entity recognition. At the same time, our results highlight that task characteristics and data domain play a central role in determining downstream performance, underscoring the importance of aligning pretraining objectives and data with target applications in Arabic NLP.

\section{Discussion}

\paragraph{Implications for Arabic Encoder Design.} Our experiments demonstrate that tokenizer design and embedding initialization are central to successful Arabic encoder modeling. The transtokenization ablation shows that introducing a new Arabic tokenizer without aligned embedding initialization leads to catastrophic degradation in masked language modeling performance. This finding reinforces the observation that Arabic’s morphological richness and lexical sparsity exacerbate tokenizer--embedding mismatches, making careful embedding initialization essential. More broadly, it suggests that future Arabic encoder models should treat tokenizer replacement as a first-class modeling decision rather than a preprocessing detail.

We also show that native long-context modeling can be effectively transferred to Arabic. AraModernBERT remains stable at sequence lengths up to 8{,}192 tokens and achieves improved intrinsic language modeling performance at extended context lengths. This result is particularly relevant for Arabic domains characterized by long-form text, such as news, legal documents, and encyclopedic content, and supports the feasibility of long-context encoders for Arabic without resorting to windowed or recurrent processing schemes.

\paragraph{Downstream Performance.} AraModernBERT transfers effectively to downstream Arabic tasks across both sentence-level classification and sequence labeling. Strong performance on semantic similarity, offensive language detection, and named entity recognition benchmarks demonstrates that gains in intrinsic modeling translate to discriminative settings. In particular, AraModernBERT performs best on larger and cleaner datasets with richer sentence-level context, such as WikiAnn for NER, suggesting alignment between its pretraining regime on long-form Arabic text and downstream data characteristics. More modest results on smaller or noisier datasets, including social media text, are consistent with prior observations for encoder models trained primarily on well-structured corpora.

\section{Conclusion}
In this work, we introduced \textbf{AraModernBERT}, an Arabic adaptation of a modern encoder architecture, and studied the role of tokenizer initialization and long-context modeling for Arabic. Our experiments show that transtokenized embedding initialization is critical for stable Arabic language modeling, leading to substantial improvements in masked language modeling performance. We further demonstrate that AraModernBERT supports native long-context modeling up to 8{,}192 tokens while remaining computationally efficient. Across downstream evaluations, AraModernBERT transfers effectively to Arabic natural language understanding and sequence labeling tasks, particularly on larger and cleaner datasets with richer sentence-level context. Overall, our findings provide practical guidance for adapting modern encoder architectures to Arabic and other Arabic-script languages.

\section*{Limitations}

This study has several limitations. While AraModernBERT supports native long-context modeling and demonstrates improved intrinsic performance at extended sequence lengths, our downstream evaluations focus on tasks that do not explicitly require long-range context at inference time. Evaluating tasks that directly benefit from long-context reasoning, such as document-level information extraction or long-form question answering, represents an important direction for future work. In addition, our experiments are limited to Arabic; although many findings are applicable to other Arabic-script languages, empirical validation on languages such as Persian, Urdu, or Kurdish remains future work. Finally, AraModernBERT is trained on approximately 100~GB of Arabic text, which, while substantial for Arabic, remains modest compared to the scale used for recent English-language encoders.

\bibliography{custom}

\end{document}